\documentclass[runningheads]{llncs}
\usepackage{graphicx}
\usepackage{amsmath,amssymb} 
\usepackage{color}
\usepackage{adjustbox}
\usepackage{multirow}
\usepackage{afterpage}

\begin{document}

\newcommand{\point}{
    \raise0.7ex\hbox{.}
    }


\pagestyle{headings}

\mainmatter

\title{Real-time Human Pose Estimation from Video with Convolutional Neural Networks} 

\titlerunning{Real-time Human Pose Estimation from Video with Convolutional Neural Networks} 

\authorrunning{Linna et al.} 

\author{Marko Linna\inst{1} \and Juho Kannala\inst{2} \and Esa Rahtu\inst{1}} 
\institute{Department of Computer Science and Engineering, University of Oulu, Finland \and Department of Computer Science, Aalto University, Finland} 

\maketitle

\begin{abstract}
In this paper, we present a method for real-time multi-person 
human pose estimation from video by utilizing convolutional 
neural networks. Our method is aimed for use case specific
applications, where good accuracy is essential and variation 
of the background and poses is limited. This enables us to use a generic 
network architecture, which is both accurate and fast. We 
divide the problem into two phases: (1) pre-training and 
(2) finetuning. In pre-training, the network is learned with 
highly diverse input data from publicly available datasets, 
while in finetuning we train with application specific data,
which we record with Kinect. Our method differs from 
most of the state-of-the-art methods in that we consider the 
whole system, including person detector, pose estimator 
and an automatic way to record application specific training 
material for finetuning. Our method is considerably faster 
than many of the state-of-the-art methods. Our method can be 
thought of as a replacement for Kinect, and it can be used 
for higher level tasks, such as gesture control, 
games, person tracking, action recognition and action tracking. 
We achieved accuracy of 96.8\% (PCK@0.2) with application 
specific data.
\end{abstract}

\section{Introduction}

Human pose estimation in the wild is a problem where humans 
yet perform better than computers. In recent years, the 
research has moved from traditional methods~\cite{felzenszwalb2008discriminatively,andriluka2009pictorial,yang2011articulated,sapp2013modec} 
towards convolutional neural networks (ConvNets, CNNs)~\cite{jain2013learning,Toshev_2014_CVPR,pfister2014deep,jain2014modeep,carreira2015human,pishchulin2015deepcut,pfister2015flowing,tompson2015efficient,lifshitz2016human,wei2016convolutional,newell2016stacked,charles2016personalizing}. 
Due to this, significant improvements in accuracy have been accomplished. 
ConvNets became popular, when AlexNet~\cite{krizhevsky2012imagenet} 
was introduced. AlexNet could classify images on different categories and it
won ILSVRC 2012\footnote{\url{http://image-net.org/challenges/LSVRC/2012}} 
competition by a significant margin to the second contestant. Since then,
several more efficient network architectures have been proposed.

GoogLeNet~\cite{szegedy2014going} introduced a new architecture, where the 
network consisted of Inception Modules. The key idea
of an Inception Module is to feed the input data simultaneously to several convolutional 
layers and then concatenate outputs of each layer into a single output. Each convolutional
layer have different filter size and they produce spatially equal sized outputs. 
Because of this, a single Inception Module can process information at various scales simultaneously,
thus leading to better performance.
In order to avoid computational blow up, Inception Modules utilize $1\times1$ convolutions
for dimension reduction. 
The main benefit of this architecture is that it allows for increasing both the depth and width of the network,
while keeping computational complexity in control. 

He et al. introduced the Residual Network (ResNet) architecture~\cite{he2015deep},
where traditional stacking convolutional layers were replaced by Residual 
Modules. In a single Residual Module, a couple of stacking 
convolutional layers are bypassed with a skip connection. The output of 
the skip connection is then added to the output of the stacking layers. 
Every convolutional layer in a Residual Module utilizes Batch Normalization 
to cope with internal covariate shift.
A typical ResNet architecture consists of a great number of stacked Residual Modules,
making the network much deeper, from tens to hundreds of layers, compared to
traditional networks. A very deep residual network is easier to optimize
than its counterpart, a plain stacking layer network. The training error is 
much lower when the depth increases, which in turn gives accuracy improvements. 
Recently, He et al. proposed an improvement to Residual Module, 
which further makes training easier and improves generalization~\cite{he2016identity}.
Zagoruyko and Komodakis argued that very deep ResNet architectures are not needed 
for state-of-the-art performance~\cite{zagoruyko2016wide}.
They decreased the depth of the network and increased the size of a Residual Module
by adding more features and convolutional layers. Currently, ResNets are state-of-the-art ConvNet models and
they have been shown to perform remarkable well both in image 
recognition~\cite{he2015deep,he2016identity,zagoruyko2016wide} and human pose estimation tasks~\cite{newell2016stacked,insafutdinov2016deepercut}.

Many state-of-the-art ConvNet human pose estimation methods uses more complex 
network architectures and they perform considerably well in unconstrained 
environments~\cite{lifshitz2016human,newell2016stacked,insafutdinov2016deepercut}, 
where large variations in 
pose, clothing, view angle and background exists. While these methods have 
high accuracy, they are usually slow considering real 
time pose estimation. Recent research~\cite{Toshev_2014_CVPR,pfister2014deep} 
shows that by using a generic ConvNet architecture, a 
competitive accuracy can be achieved, while still 
maintaining a short forward pass time. This is the main motivation
of our research. With our method, we don't aim for overall human pose 
estimation in diverse input data, but rather target to specific use 
cases where high accuracy and speed are required. In such cases, 
the problem is different, because the environment is usually constrained, 
persons are in close proximity of the camera and poses are restricted. 
Possible application for our method are, for instance, gesture control 
systems and games. 

Our method is a multi-person human pose estimation system, targeted for 
use case specific applications. In order to support multiple people, we use a person detector, which
gives locations and scales of the persons in the target image. This 
brings our method towards the practice, since person location and 
scale are not expected to be known, which is the case with many 
state-of-the-art methods ~\cite{lifshitz2016human,wei2016convolutional,newell2016stacked}. 
We use a generic ConvNet architecture, with eight layers. The key idea of 
our method is to pre-train the network with highly diverse input 
data and then finetune it with use case specific data. We show that 
competitive accuracy can be achieved in application specific pose estimation,
while operating in real-time. Our method can be used for higher 
level tasks, for example, gesture control, gaming, action recognition and
action tracking.

The main contributions of our method are: 
(1) working replacement for Kinect~\cite{shotton2013real} by using a fast and accurate
pose estimation network together with a state-of-the-art person detector,
(2) utilization of Kinect for automatic training data generation, making 
it easy to generate large amount of annotated training data, 
(3) utilization of person detector to crop person centered images in both 
training and testing, thus enabling multi-person pose estimation in real 
world images,
(4) ability to learn from heterogeneous training data, where the set of joints is 
not the same in all the training samples, thus enabling to use 
more varied datasets in training.

\section{Related Work}

Jain et al.~\cite{jain2013learning} demonstrated that ConvNet based human pose estimation can 
meet the performance, and in many cases outperform, traditional methods, particularly 
deformable part models~\cite{felzenszwalb2008discriminatively} and multimodal 
decomposable models~\cite{sapp2013modec}. Their network architecture 
consisted of three convolutional layers, followed by three fully connected layers.
Pooling was applied after the first two convolutional layers. They trained the network
for each body part (e.g. wrist, shoulder, head) separately. Each network was applied as sliding windows to overlapping
regions of the input image. A window of pixels was mapped to a single binary output: 
the presence or absence of that body part. This made possible to use much smaller 
network, at the expense of having to maintain a separate set of parameters for each 
body part.

Another application to human pose estimation was presented by Toshev and 
Szegedy~\cite{Toshev_2014_CVPR}. Their network architecture was similar to AlexNet~\cite{krizhevsky2012imagenet},
but the last layer was replaced by a regression layer, which output joint coordinates. 
In addition to this, they trained a cascade of pose regression networks. The cascade 
started off by estimating an initial pose. Then at subsequent stages, additional regression
networks were trained to predict a transition of the joint locations from previous stage to 
the true location. Thus, each subsequent stage refined the currently predicted pose.
Similar idea is applied in more recent work by Carreira et al.~\cite{carreira2015human}.

A video based human pose estimation method was introduced by Pfister et al.~\cite{pfister2014deep}.
Their method utilized the temporal information available in constrained gesture videos. This was achieved
by training the network with multiple frames so that the frames were inserted into the 
separate color channels of the input. For example, with three input frames, the number 
of color channels would be nine. The network architecture was similar to AlexNet, having five 
convolutional layers, followed by three fully connected layers, from which the last
one was a regression layer. Pooling was done after the first, second and fifth convolutional
layer. However, there were some differences compared to the previous architectures. Some 
of the convolutional layers were much deeper and pooling was non-overlapping, when in most
of the previous architectures it was overlapping. The network produced significantly better 
pose predictions on constrained gesture videos than the previous work. For this reason, we
base our method to this architecture.

Since introducing the first ConvNet method for human pose estimation~\cite{jain2013learning},
a number of related methods have been proposed.
While most of the methods focus on estimating poses in isolated 
still images, only few concentrate on pose estimation in videos. 
Utilizing of the temporal information of subsequent frames of a video
may be a valuable cue when estimating keypoint locations. To this 
subject, dense optical flow has been used successfully in several 
works. Jain et~al.~\cite{jain2014modeep} use it to create motion 
feature images, which are fed to ConvNet together with corresponding 
RGB frames. In addition, optical flow has been used in 
~\cite{pfister2015flowing,charles2016personalizing} to warp keypoint 
heatmaps of neighboring frames in order to reinforce the confidence of 
the current frame.

Recently, Newell et al.~\cite{newell2016stacked} introduced a new ConvNet 
architecture for human pose estimation, which achieved state-of-the-art results 
on the FLIC~\cite{sapp2013modec} and MPII~\cite{andriluka14cvpr} benchmarks
outperforming all recent methods. Their network architecture benefit 
from recently introduced ResNets~\cite{he2015deep,he2016identity}, convolution-deconvolution 
architectures~\cite{noh2015learning} and 
intermediate supervision~\cite{wei2016convolutional}. The core of the architecture
is the Hourglass Module, which implements bottom-up, top-down architecture, making 
possible to better process features across different scales.
The Hourglass Network consists of two stacked Hourglass Modules and it
outputs heatmaps in two stages, where the network predicts the probability 
of each joint’s presence at every pixel. The first stage outputs initial predictions,
while the second outputs final predictions. In training with intermediate supervision, 
the loss is applied for both predictions separately using the same ground truth.

Majority of the recent human pose estimation methods focus on a
single person case, where the approximate location of a person 
is expected to be known. However, there has been also some 
research in multi-person case, where the close proximity of different
persons causes challenges to pose estimation. Pishchulin et 
al.~\cite{pishchulin2015deepcut} uses body part detection and pose 
estimation jointly to draw a conclusion about the number of persons 
in an image, identify occluded body parts and disambiguate body parts 
between people in close proximity of each other. Their method differs 
from previous work in that they don't use separate person detection and 
pose estimation steps, but instead solve both problems together.
The main idea of their method is to use ConvNet and graphical model jointly.
The method starts by detecting body part candidates (e.g. potential head, shoulder or knee) 
by utilizing an altered version of Fast R-CNN~\cite{girshick2015fast}.
Then the body part candidates are used to form a graph, where every distinct 
body part is connected to all other body parts by a pairwise term. A pairwise
term is used to generate a cost or reward to be paid by all feasible solutions 
of the pose estimation problem for which the both body parts belong to the same 
person. The pose estimation problem is regarded as an Integer Linear Programming 
(ILP) that minimizes over the set of feasible solutions. Additional costs, 
variables and constraints ensure that feasible solutions
unambiguously selects and classifies body part candidates as body part classes,
and that body part candidates are clustered into distinct people.

Our method differs from many recent methods in that we use a person
detector to locate persons from an image and then apply pose
estimation for each person individually. Our method can be
considered as an end-to-end system, as we include all the
required steps for pose estimation from arbitrary source image.

\section{Method}

\begin{figure}[t]
  \begin{center}
    \includegraphics*[width=0.9\textwidth]{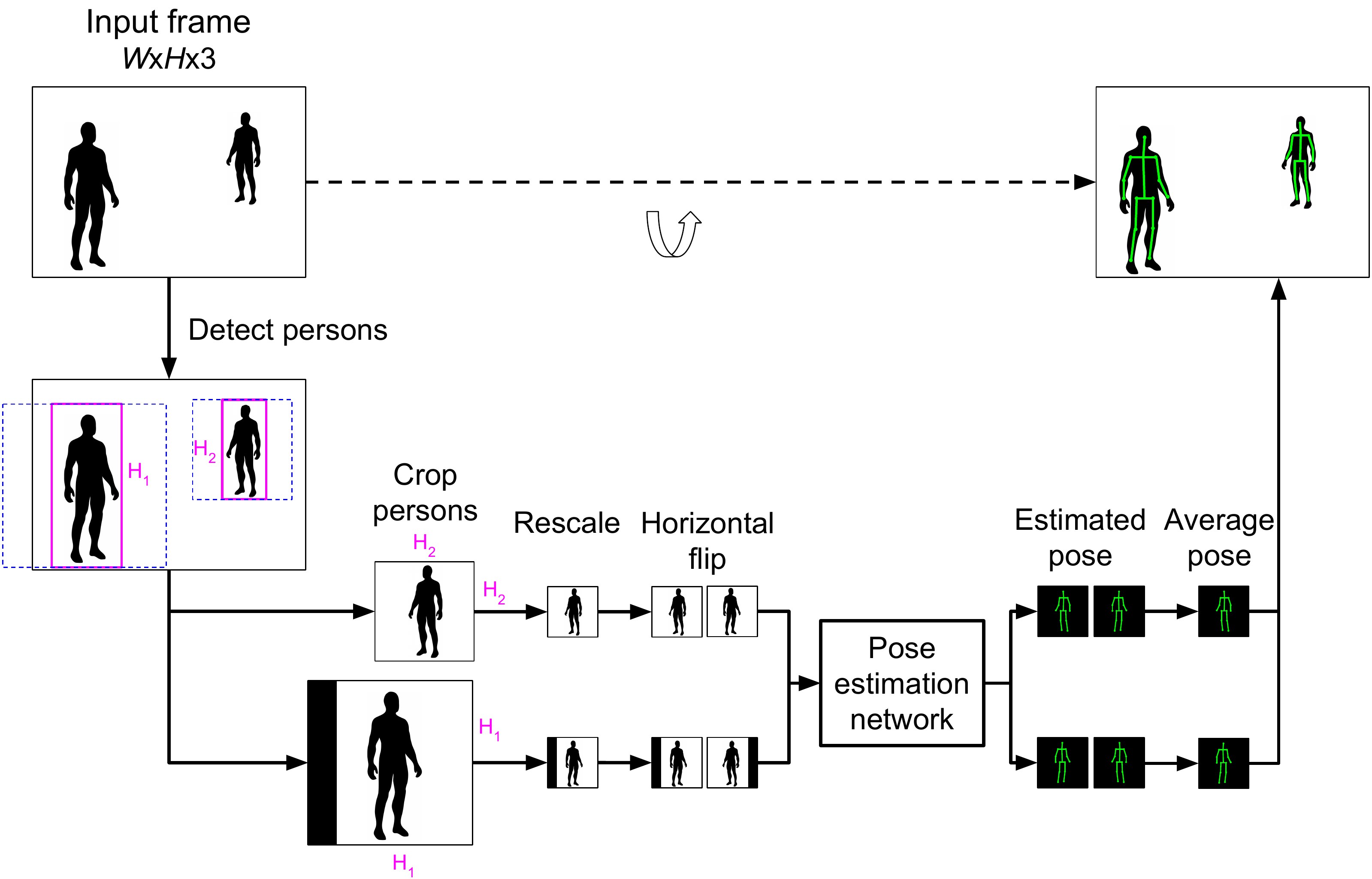}
  \end{center}
  \caption{The pose estimation process for a single input frame.  
  Zero padding is added for cropped person images on regions outside the image borders.
  Person images are rescaled to size $224\times224$ before feeding them to the network.}
  \label{fig:pose_estimation_process}
\end{figure}

Our method is targeted for video inputs.
The rough steps for a single video frame in testing are:
(1) detect persons,
(2) crop person centered images,
(3) feedforward person images to the pose estimation network.
We use a separate object detector~\cite{ren15fasterrcnn} to solve person 
bounding boxes from the input frame. The pose estimation is done for 
each person individually. As a result of the pose estimation, our 
network outputs locations of body keypoints. 
The pose estimation process is described in Figure~\ref{fig:pose_estimation_process}.

We pre-train our network 
by using data from multiple publicly available datasets, thus 
offering good initialization values for finetuning. We evaluate 
pre-training and finetuning separately. For the evaluation of the 
finetuning, we use data recorded with Kinect. As for ConvNet framework, 
we use Caffe~\cite{jia2014caffe} with small modifications. 

\subsection{Person Detection}

Our method utilize Faster R-CNN (F-RCNN)~\cite{ren15fasterrcnn} to 
detect persons from training and testing images. The forward pass time 
of the F-RCNN is 60ms or 200ms, depending on the used network. We use
the slower and more accurate model. 

We noticed that sometimes F-RCNN 
gives false positives. This is not a problem in training, since we use
both the ground truth and the F-RCNN together to crop the training 
image. But in testing, the pose estimation is also performed for false 
positives. However, most likely these false positives could be filtered,
especially with use case specific images, by adjusting the parameters of the 
F-RCNN. In the evaluation, we use also the ground truth to decide if 
the frame has a person or not, so it is guaranteed that all the evaluation
frames contain a person. Apart from this, we ran the F-RCNN for the 
original finetuning evaluation frames, where the ground truth was not yet used for
the frame selection. This resulted in false positive rate of 2.86\% and 
false negative rate of 0.65\%. In all of the original evaluation frames, there  
is one fully visible person making gestures in constrained environment.
Person detection was considered false if the resulted bounding box did 
not contain a person, or if it had partially visible person on the edges
of the bounding box. In other words, if the intersection-over-union (IoU)
ratio between the detection and the ground truth was 0.5 or less.

\subsection{Data Augmentation}

The F-RCNN person detector is applied for each training image. For each 
detected person, the IoU between the detected 
person bounding box and the expanded ground truth bounding box is 
calculated. The expanded ground truth person box is the tightest bounding 
box, including all the joints, expanded by a factor of 1.2. The person 
box having the biggest IoU is selected as the 
best choice. Based on the best IoU, the training image is augmented by using 
either of the person bounding boxes, or both (see Table~\ref{table:IoU}).

\begin{table}[b]
\setlength{\tabcolsep}{4pt}
\begin{center}
\scriptsize 
\begin{tabular}{lcc}
\hline
\noalign{\smallskip}
                                    & \multicolumn{2}{c}{Person box type used in augmentation} \\
\\
Overlapping ratio                   & F-RCNN                & Ground truth                     \\ 
\noalign{\smallskip}
\hline
\noalign{\smallskip}
$\textup{IoU} > 0.7$                & X                     &                                  \\
$\textup{IoU} < 0.5$                &                       & X                                \\ 
$0.5 \geq \textup{IoU} \leq 0.7$    & X                     & X                                \\ 
\noalign{\smallskip}
\hline
\end{tabular}
\end{center}
\caption{The relation between the person box overlapping ratio and the 
data augmentation.}
\label{table:IoU}
\end{table}

In practice, this means that if the detected person box is near to the 
ground truth expanded person box, only the former is used to crop the 
person image. And if the detected person box is far from the 
ground truth expanded person box, only the latter is used to crop the 
person. And in between of these, both person boxes are used to crop 
the person, resulting in two training images, where both have small 
differences in translation and scale. The shortest side of a person 
box is expanded to equal the longest side, resulting a square crop 
area, defining the person image used in training. Zero padding is 
added where needed. A single cropped person image is rescaled to 
size $224\times224$ before feeding it to the network.

\begin{figure}[t]
  \begin{center}
    \includegraphics*[width=0.7\textwidth]{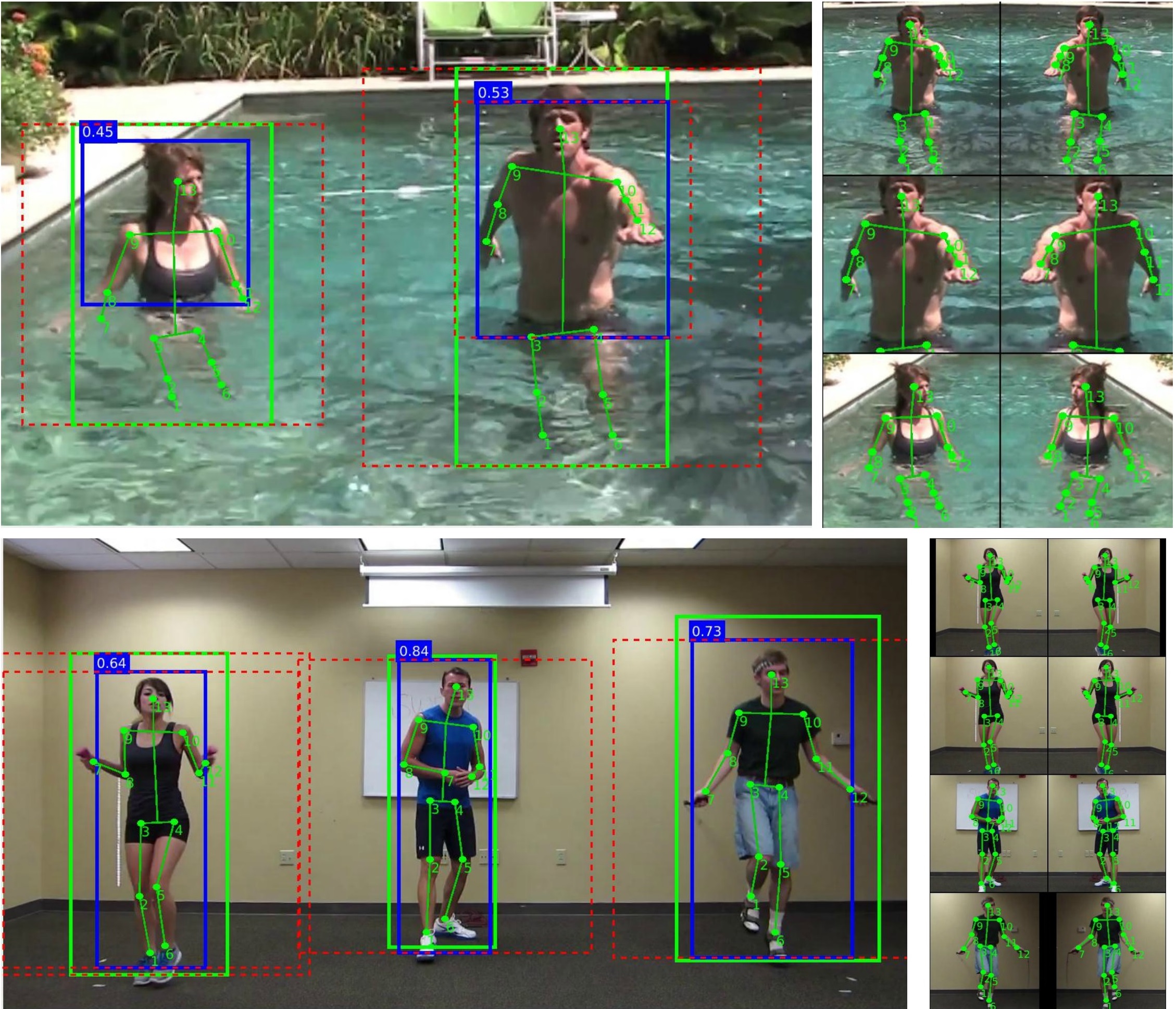}
  \end{center}
  \caption{Visualization of data augmentation with two random samples from the
  MPII Human Pose dataset~\cite{andriluka14cvpr}. On the left column are the image samples and 
  on the right the cropped person centered images for each image sample. 
  Ground truth (the points in the skeleton) and the ground truth 
  expanded bounding boxes are in green. Detected person boxes are in 
  blue and the value on the top-left corner of the box is the overlapping 
  ratio (IoU) between the ground truth expanded box and the detected person 
  box. The crop area is in red (dashed square) and it is expanded 
  according to the longest side of the person bounding box.  
  The top image sample shows well that when person detection fails to
  capture the whole person, the ground truth is used to crop the
  person image. Otherwise, the goal is to use the detected person box.}
  \label{fig:data_aug_vis}
\end{figure}

In addition to aforementioned, a training image is augmented by 
doing a horizontal flip, giving double version of the image. All in 
all, a single person image from a source dataset can result in 
either two or four augmented person centered training images. 
Data augmentation is visualized in Figure~\ref{fig:data_aug_vis}.

\subsection{Pre-training}

We pre-train the model from scratch by using several publicly 
available datasets (see Table~\ref{table:datasets}). The number of
annotated joints varies between the datasets. The MPII Human Pose
~\cite{andriluka14cvpr}, Fashion Pose~\cite{DGLG13} and Leeds Sports 
Pose~\cite{Johnson10} have full body annotations, while the FLIC
~\cite{sapp2013modec} and BBC Pose~\cite{Charles13a} have only upper body
annotated. Since we use a single point for the head, and because the 
MPII Human Pose and Leeds Sports Pose have annotations for the neck 
and head top, we take the center point of these and use it as a head 
point. 

\begin{table}[t]
\setlength{\tabcolsep}{4pt}
\begin{center}
\scriptsize 
\begin{tabular}{lllllll}
\hline
\noalign{\smallskip}
                                       &  & \multicolumn{3}{c}{\begin{tabular}[c]{@{}c@{}}Person boxes we use\\from the dataset\end{tabular}} & \multicolumn{2}{c}{\begin{tabular}[c]{@{}c@{}}Person boxes we use for\\pre-training and validation\end{tabular}} \\
                                       & \multirow{2}{*}{\begin{tabular}[c]{@{}l@{}}Annotated\\points\end{tabular}} \\
Dataset                                &      & Train                            & Test                           & Total                           & Train (aug.)                                            & Validation                                            \\ 
\noalign{\smallskip}
\hline
\noalign{\smallskip}
MPII Human Pose~\cite{andriluka14cvpr} & 1-16 & 28821                            & 0                              & 28821                           & 71018                                                   & 1160                                                  \\
Fashion Pose~\cite{DGLG13}             & 13   & 6530                             & 765                            & 7295                            & 14538                                                   & 694                                                   \\
Leeds Sports Pose~\cite{Johnson10}     & 14   & 1000                             & 1000                           & 2000                            & 5074                                                    & 146                                                   \\
FLIC~\cite{sapp2013modec}              & 11   & 3987                             & 1016                           & 5003                            & 14780                                                   & 0                                                     \\
BBC Pose~\cite{Charles13a}             & 7    & 0                                & 2000                           & 2000                            & 6764                                                    & 0                                                     \\ 
\noalign{\smallskip}
\hline
\noalign{\smallskip}
                                       &      & 40338                            & 4781                           & 45119                           & 112174                                                  & 2000                                                  \\ 
\noalign{\smallskip}
\hline
\end{tabular}
\end{center}
\caption{Overview of used datasets in pre-training. Only the training 
set of the MPII Human Pose is used, because the annotations are not 
available for the test set. In the BBC Pose, the training set is 
annotated semi-automatically~\cite{Buehler11}, while the test set is 
manually annotated. We use only manually annotated data from the BBC 
pose. We use data augmentation to expand the number of training images.}
\label{table:datasets}
\end{table}

\begin{figure}[p]
\begin{center}
\includegraphics*[width=0.8\textwidth]{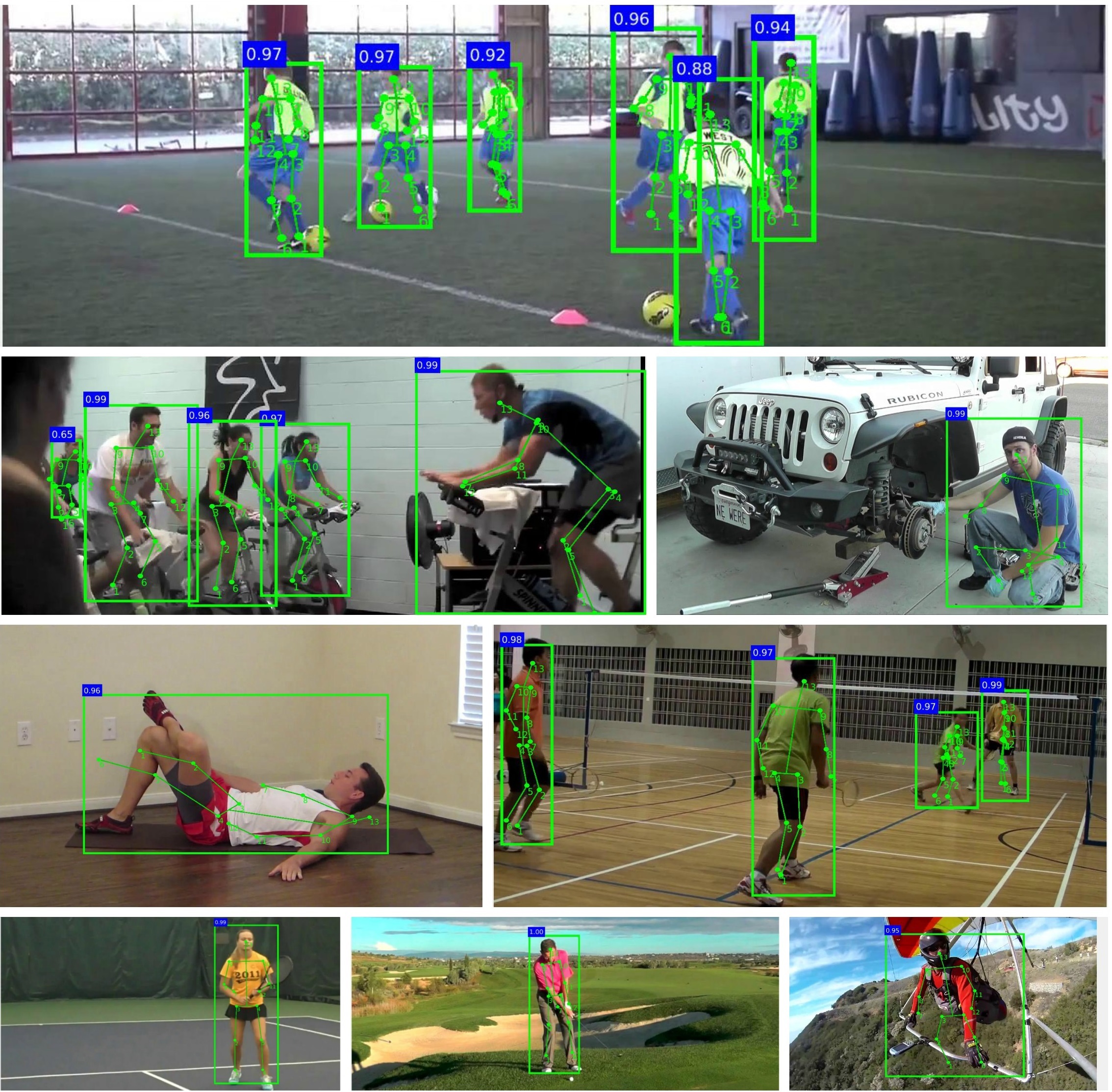}
\caption{Example pose estimations with the pre-trained network. 
Samples are taken randomly from the testing set of the MPII Human Pose 
dataset. The green bounding boxes are the results of person detection and the
number on the top-left corner is the probability of a box containing a person.}
\label{fig:pretrain_estimations}
\vspace{\floatsep}
\includegraphics*[width=0.8\textwidth]{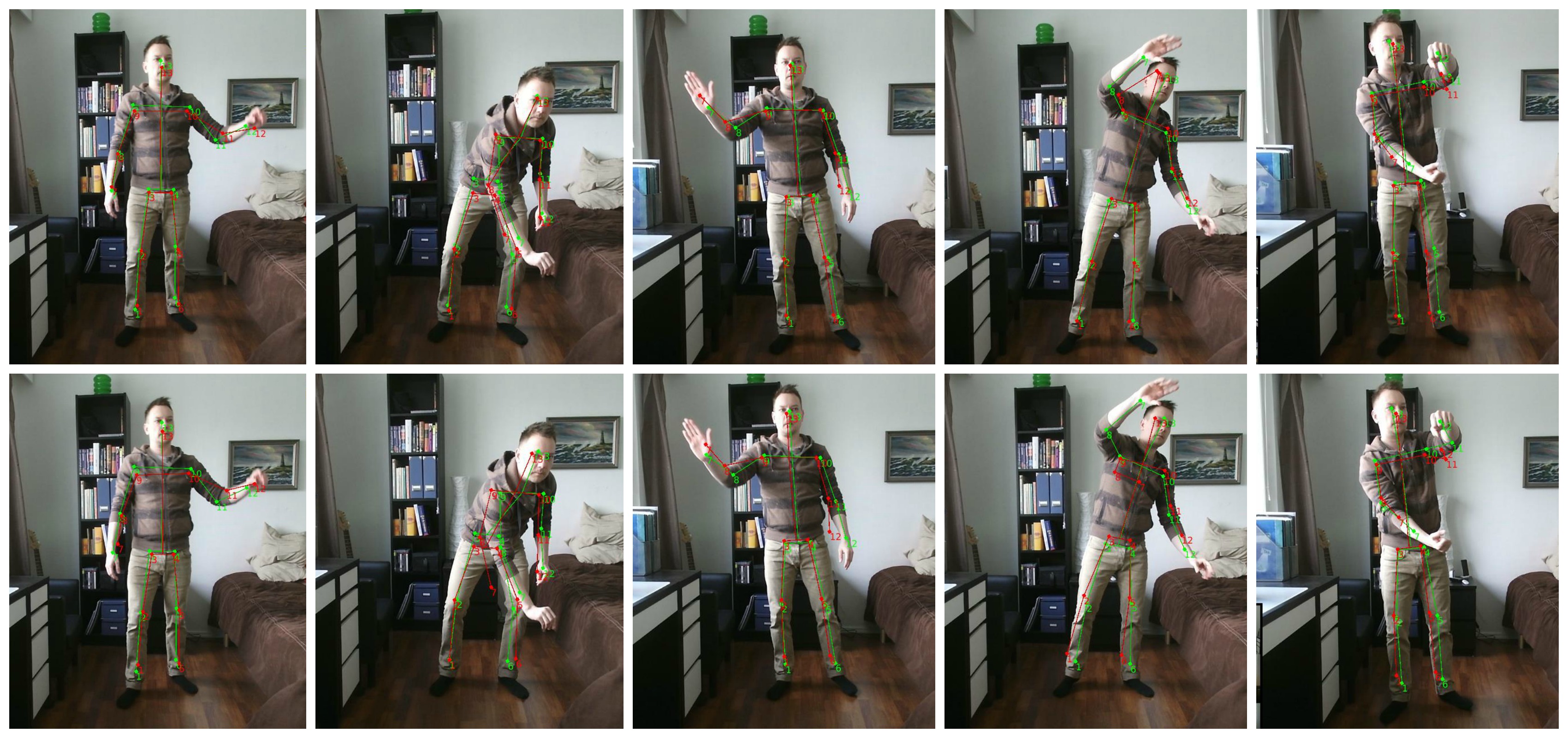}
\caption{Example pose estimations with the finetuned network.
Predictions are in red and Kinect ground truth in green.
On the columns are five different frames from the evaluation data. 
The first row shows results of the full finetuning (experiment 3) and
the second row shows results of the phase 1 (experiment 1). 
Experiments are explained later in Section~\ref{sec:evaluation}.
Full videos are available at \url{https://youtu.be/qjD9NBEHapY}
and \url{https://youtu.be/e-P5SYL-Aqw}.}
\label{fig:finetune_samples}
\end{center}
\end{figure}

As we aim to study that whether additional partially annotated training 
data brings improvement over using only fully annotated samples, our 
validation samples should be fully annotated. Thus, we put all
the fully annotated (13 joints) person images to a
single pool and sample 2000 images randomly for validation. The validation
images are then removed from the pool. Next, we put all the partially annotated 
images to the same pool so that it eventually contains person images with 
heterogeneous set of annotated joints. Then we use the pool 
in training. The purpose of the pre-trained model is to offer a good 
weight initialization values for finetuning. Pre-training takes 23 
hours on three NVIDIA Tesla K80 GPUs.
Figure~\ref{fig:pretrain_estimations} contains example pose estimations
with the pre-trained network. 

\subsection{Finetuning}

The purpose of the finetuning is to adapt the pre-trained model for the 
particular use case. For instance a gesture control system or a game. 
The pre-trained model alone is not a good enough pose estimator for our 
use cases, because the shallow network we use lacks the
capacity to perform well with highly diverse training data. More 
complicated network architectures, such as ~\cite{newell2016stacked,lifshitz2016human} 
would certainly give better results, but then the speed gain achieved 
with shallow network architecture would most likely be lost. 

In finetuning, the pre-trained model is used for weight initialization.
When the network is finetuned with use case specific data, for example 
to estimate poses in gesture control system, the training data 
is most likely consistent. This is a good thing when thinking of 
accuracy. Even a shallow network can produce very good 
estimations, if the training data is limited to particular
use case. Using more complicated, and potentially slower, network 
architectures in these situations is therefore not necessary. 
We use Kinect in our experiments to produce annotations for the finetuning data, 
but alternative methods can be considered as well. Figure~\ref{fig:finetune_samples} 
contains example pose estimations with the finetuning evaluation data.

\subsection{Network Architecture}

\begin{figure}[t]
\centering
\includegraphics[width=1\textwidth]{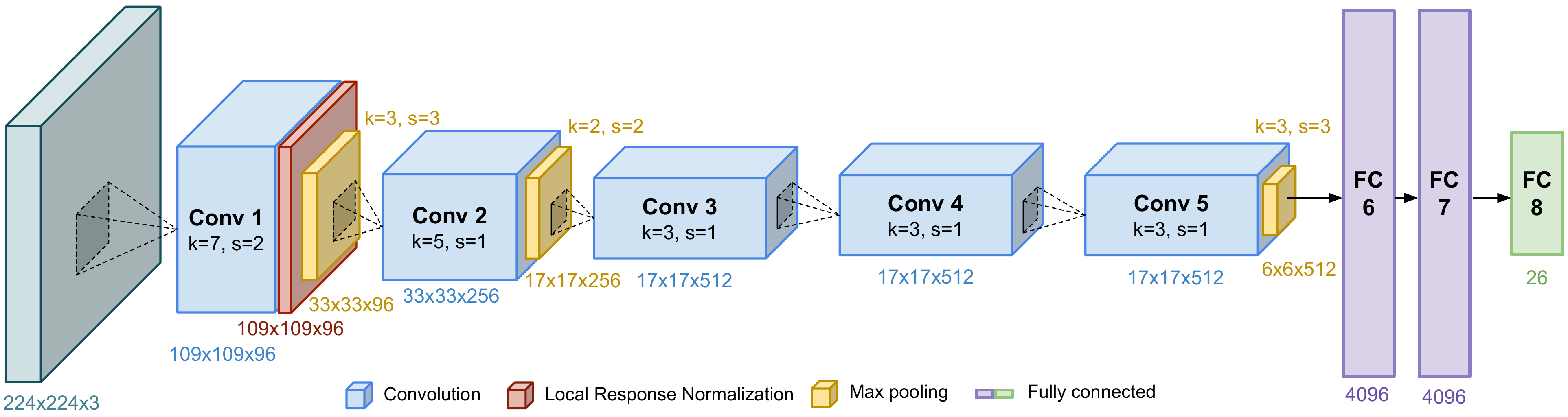}\\
\caption{The architecture of the pose estimation network. Letters k and s 
means kernel size and stride.}
\label{fig:net_arch}
\end{figure}

Our method utilizes generic ConvNet architecture, having five convolutional
layers followed by three fully connected layers, from which the last layer 
is regression layer (see Figure~\ref{fig:net_arch}). The regression layer 
produces $(x, y)$ position estimates for human body joints. More closely, 
one estimation for head, six for arms and six for legs, a total of 13 position estimations. The network input size is 
$224\times224\times3$. The network does not utilize any spatiotemporal 
information, but treats all training images individually. We use generic 
ConvNet architecture, because it has shown to perform well in human pose 
regression tasks~\cite{Toshev_2014_CVPR,pfister2014deep}. The forward 
pass time of the network is 16ms on Nvidia GTX Titan GPU, which makes it 
highly capable for real-time tasks.

\subsection{Training Details}

In model optimization, the network weights are updated using batched 
stochastic gradient descent (SGD) with momentum set to 0.9. 
In pre-training, where the network is trained from scratch, the learning 
rate is set to $10^{-2}$, weights are initialized randomly using Xavier 
algorithm~\cite{Glorot10understandingthe} and biases are set to zero. 
In finetuning, the learning rate is set to $10^{-3}$. The loss function 
we use in optimization, penalizes the distance between predictions and 
ground truth. We use weighted Euclidean (L2) loss

\begin{equation} 
E=\dfrac{1}{2N} \sum_{i=1}^N w_i\left \| x_i^{gt}-x_i^{pred} \right \| _2^2
\end{equation} 

where vectors $w$, $x^{gt}$ and $x^{pred}$ holds joint coordinates and 
weights in form of $(x_1$, $y_1$, $x_2$, $y_2$, $...$, $x_{13}$, 
$y_{13})$. Weight $w_i$ is set to zero if the ground truth of the joint 
coordinate $x_i^{gt}$ is not available. Otherwise it is set to one. 
This way only the annotated joints contribute to the loss. 
This enables training the network using datasets having only the
upper body annotations, along with datasets having full body annotations. 
Ability to utilize heterogeneous training data, where the set of joints 
is not the same in all training samples, potentially leads to better 
performance as more training data can be used. 

As for comparison, we train the pre-trained model also without using the
weighted Euclidean loss. In this case, we use only images with fully  
annotated joint positions (13 joints), so that the training data is 
homogenous regarding to joint annotations. Doing this reduces the size of 
the training data from 112174 to 66598 images. The average joint prediction 
error with heterogeneous and homogenous data are 15.7 and 16.6 pixels on
$224\times224$ images. With heterogeneous data, there is about 5\% 
improvement on prediction error.

In batched SGD, we use batch size of 256. Each iteration selects images 
for the batch randomly from the full training set. A training image 
contains roughly centered person of which joints are annotated. The
training images are resized to $224\times224$ before feeding to 
the network. Mean pixel value of 127 is reduced from every pixel 
component and the pixel components are normalized to range [-1, 1]. 
Joint annotations are normalized to range [0, 1], according to the 
cropped person centered image. 

\subsection{Testing Details}

The person detector is applied for an image from which poses are to be 
estimated. Person images are cropped based on detections as described earlier. 
In addition, for each person image, a 
horizontally flipped double is created. Both the original and the 
doubled person images are fed to the network. The final joint prediction 
vector is average of the estimations of these two (the predictions of 
the doubled image are flipped so that they correspond predictions of the 
original image). By doing this, a small gain in accuracy is achieved.

\section{Evaluation}
\label{sec:evaluation}

We evaluate pre-training and finetuning with the percentage of correct 
keypoints (PCK) metric~\cite{sapp2013modec}, where the joint location 
estimate is considered correct, if its L2 distance to the ground truth 
is at most 20\% of the torso length. The torso length is 
the L2 distance between the right shoulder and the left hip.

We use 2000 randomly taken samples for the evaluation of the pre-training. 
For finetuning, we record data with Kinect for Windows v2 (see Table
~\ref{table:finetuning_data}). We use the joint estimates produced 
by Kinect as a ground truth. We made sure that the data was recorded 
in a such way, that the error in the joint estimations is minimal. Practically 
this means good lightning conditions, no extremely rapid movements and 
no major body part occlusions. The gestures performed in the data tries 
to mimic different gesture control events, where the hands are used for 
tasks like object selection, moving, rotating and zooming, in addition to 
hand drawing and wheel steering. 

\begin{table}[t]
\setlength{\tabcolsep}{4pt}
\begin{center}
\scriptsize 
\begin{tabular}{ll}
\hline
\noalign{\smallskip}
Clothing & Frames \\ 
\noalign{\smallskip}
\hline
\noalign{\smallskip}
1        & 27222  \\
2        & 18760  \\
3        & 20244  \\
4        & 20726  \\
5        & 10560  \\
6        & 11666  \\
7        & 10136  \\ 
\noalign{\smallskip}
\hline
\noalign{\smallskip}
         & 119314 \\ 
\noalign{\smallskip}
\hline
\end{tabular}
\end{center}
\caption{Kinect recorded finetuning data for training. 
All the frames have similar background, person and gestures, 
but clothing differs. For the evaluation, we additionally record 4000 
frames, which have identical clothing (clothing number 1).}
\label{table:finetuning_data}
\end{table}

\begin{table}[t]
\setlength{\tabcolsep}{4pt}
\begin{center}
\scriptsize 
\begin{tabular}{llll}
\hline
\noalign{\smallskip}
\# & Name & \begin{tabular}[t]{@{}l@{}}Initialization\\network\end{tabular} & \begin{tabular}[t]{@{}l@{}}Clothing in\\training frames\end{tabular} \\ 
\noalign{\smallskip}
\hline
\noalign{\smallskip}
1  & phase 1 & pre-train              & 2,3,4,5,6,7                 \\
2  & phase 2 & phase 1                & 1                           \\
3  & full    & pre-train              & 1,2,3,4,5,6,7               \\ 
\noalign{\smallskip}
\hline
\end{tabular}
\end{center}
\caption{Finetuning experiments. The training data have (1) different clothing
from the testing data in every frame, (2) the same clothing as the 
testing data in every frame, (3) the same clothing as the testing data 
in some of the frames. In phase 2, the finetuning is done over 
already finetuned network of the phase 1. Otherwise it is done over the 
pre-trained network.}
\label{table:finetuning_phases}
\end{table}

For the evaluation of the finetuning, we record additional 4000 frames 
with identical clothing . We do three finetuning experiments,
using different set of training frames in each case (See Table~\ref{table:finetuning_phases}). 
The experiments 1 and 2 together uses the same training frames as the
experiment 3. Basically, the experiment 3 is the same as the experiments 
1 and 2 performed consecutively. The purpose of this divide is to
see the effect of using the same/different clothing between the training and
testing data. The experiment 1 express more of the ability of generalization 
(for all people) while the experiments 2 and 3 of specificity (for certain people).

\afterpage{
\begin{figure}[t]
\centering
\adjustbox{trim={.1\width} {.09\height} {0.06\width} {.08\height},clip}%
  {\includegraphics[width=14.4cm]{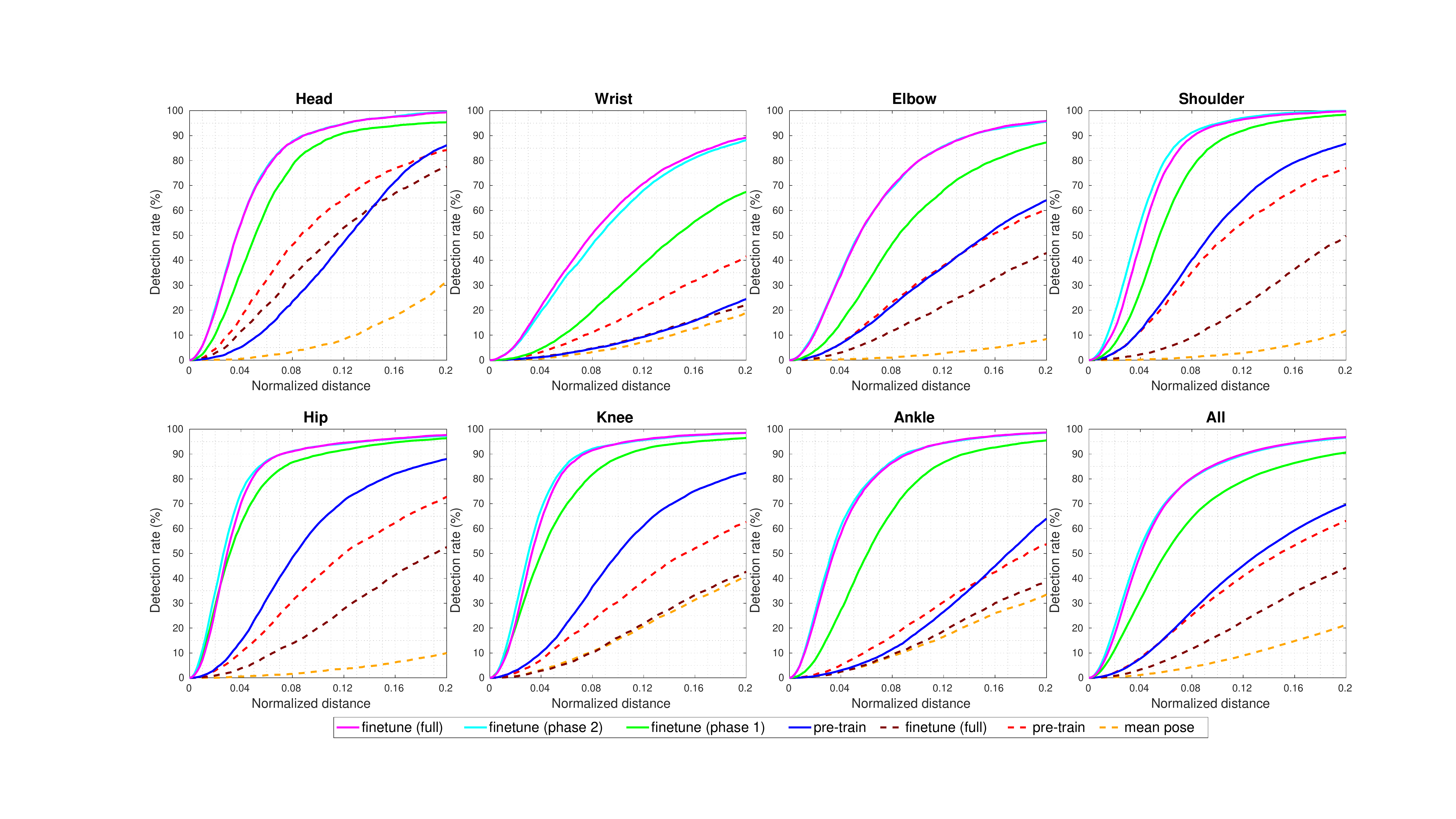}}
\caption{The results of pose estimation (PCK@0.2). The dashed lines uses pre-train
validation samples (2000 images) in testing, while the solid lines use 
finetuning validation samples (4000 frames). To put it other way, the 
dashed lines represent the accuracy of generalization, while the solid lines
represent the use case specific accuracy. The label indicates which
network is used in testing.}
\label{fig:eval_accuracy}
\end{figure}
\setlength{\tabcolsep}{3.5pt}
\begin{table}[t!]
\setlength{\tabcolsep}{4pt}
\begin{center}
\scriptsize 
\begin{tabular}{llllllllll}
\hline
\noalign{\smallskip}
Network            & Head            & Wrist           & Elbow           & Shoulder        & Hip             & Knee            & Ankle           & All             \\ 
\noalign{\smallskip}
\hline
\noalign{\smallskip}
Mean pose          & 31.1            & 18.9            & 8.5             & 11.8            & 10.0            & 40.8            &          33.5   &          21.4   \\
Pre-train          & 84.2            & 41.6            & 60.5            & 76.9            & 72.8            & 62.6            &          53.7   &          63.1   \\
Finetune (full)    & 77.5            & 22.2            & 42.9            & 49.8            & 52.5            & 42.6            &          38.6   &          44.2   \\ 
\noalign{\smallskip}
\hline
\noalign{\smallskip}
Pre-train          & 86.1            & 24.5            & 64.1            & 86.8            & 88.0            & 82.5            &          64.0   &          69.6   \\
Finetune (phase 1) & 95.3            & 67.4            & 87.3            & 98.4            & 96.3            & 96.4            &          95.5   &          90.6   \\
Finetune (phase 2) & \textbf{99.6  } & 88.1            & 95.6            & \textbf{99.9  } & 97.3            & \textbf{98.5  } &          98.5   &          96.6   \\
Finetune (full)    & 99.3            & \textbf{89.2  } & \textbf{95.9  } & 99.7            & \textbf{97.6  } & \textbf{98.5  } & \textbf{98.6  } & \textbf{96.8  } \\ 
\noalign{\smallskip}
\hline
\end{tabular}
\end{center}
\caption{The results of pose estimation (PCK@0.2). The first three cases 
uses pre-train validation samples (2000 images) in testing, 
while other models use finetuning validation samples (4000 frames).}
\label{table:eval_results}
\end{table}
} 

The results are displayed in Figure~\ref{fig:eval_accuracy} and 
Table~\ref{table:eval_results}. In full finetuning (experiment 3), with the
use case specific data, the accuracy of 96.8\% is achieved. 
In finetuning phase 1 (experiment 1), where no same clothing occurs between the 
training and testing data, the accuracy is 90.6\%.
However, if we look at the accuracy of wrist (pre-train: 24.5\%, phase 1: 67.4\%, full: 89.2\%), 
which is the most challenging body joint to estimate, but perhaps also the 
most important one considering a gesture control system, we can see that 
additional case specific training data can significantly improve the 
accuracy and make the system usable in practice.
This originates partially 
from the finetuning data, where the wrist location variation is biggest. 
We believe, that if more training data would be used, and perhaps a
better data augmentation, a better wrist accuracy could be achieved with 
the current network architecture. After all, the wrist accuracy is still 
decent, making our method useful for many use cases.

The results indicate that a trade-off between generalization and 
specificity exists between pre-training and finetuning. This can be 
seen by comparing accuracies between the pre-trained and finetuned 
networks, first with the pre-train validation samples and then with 
the finetuning validation samples. The pre-train validation samples express 
the case of generalization as they contain a large variation of persons
and poses in unconstrained environment. On the contrary, the finetuning 
validation samples reflects the case of specificity as they have
restricted poses in constrained environment. After the full finetuning, 
the accuracy on the pre-train validation set drops from 63.1\% to 44.2\% 
(light red and dark red curves in Figure~\ref{fig:eval_accuracy}), 
while in the same time, the use case specific accuracy increases 
from 69.6\% to 96.8\% (blue and magenta curves). In certain cases, 
the loss in generalization is acceptable, if at the same time, gain 
in specificity is achieved. One example of a such case is a gesture control system set up in a factory, 
where all the persons wear identical clothing. 
Most importantly, while generic person detection in highly varying poses 
and contexts is an important and challenging problem, our results show 
that in some use cases the state-of-the art for the generic problem may 
produce inferior results compared to a simpler approach which has been 
specifically trained for the problem at hand.

\section{Conclusion}

We introduced a real-time ConvNet based system for human pose estimation and 
achieved accuracy of 96.8\% (PCK@0.2) by finetuning the network
for specific use case. Our method can be thought of as a replacement 
for Kinect, and it can be used in various tasks, like gesture control, 
gaming, person tracking, action recognition and action tracking. Our method supports 
heterogeneous training data, where the set of joints is not the same in 
all the training samples, thus enabling utilization of different datasets in
training. The use of a separate person detector brings our method 
towards the practice, where the person locations in the input images are
not expected to be known. In addition, we demonstrated an automatic and easy way to 
create large amounts of annotated training data by using Kinect. 
The network forward time of our method is 16ms, without the person
detector and with the person detector, either 60+16=76ms or 
200+16=216ms.

As for future work, there are several things that could be considered
in order to get better accuracy. One option would be to use 
current network as a coarse estimator and use another network for 
refining the pose estimation. In addition, as our method is targeted
for video inputs, the utilization of the spatiotemporal data would
most likely give accuracy boost. The network forward time of the 
person detector is relatively slow compared to the pose estimation 
network (16ms vs. 60ms/200ms). While the person detector works well
with diverse input data, perhaps, with most pose estimation use cases, 
that is not necessary. By using more restricted and possibly faster 
person detector, a good enough performance in more constrained 
environments could be most likely achieved. Also, with 
ConvNets, generally, holds that if more data used in training, 
the better performance gained. Hence, the use of more advanced data 
augmentation methods, such as ~\cite{pishchulin2012articulated}, 
especially in the finetuning, would most probably lead to better 
accuracy. Advanced data augmentation could, for example, change 
colors of the clothes, adjust limb poses and change backgrounds.

\bibliographystyle{splncs}
\bibliography{accv2016finalpaper}



\end{document}